# Interpretable Prostate Cancer Detection using a Small Cohort of MRI Images

Vahid Monfared[1], Mohammad Hadi Gharib[2,3], Ali Sabri[4], Maryam Shahali[5], Farid Rashidi[6], Amit Mehta[7,8],

Reza Rawassizadeh[1,9]

1. Department of Computer Science, Boston University Metropolitan College, Boston, MA, USA (vahidm@bu.edu, rezar@bu.edu)

2. MD, Clinical Fellow, Department of Radiology, McMaster University, Ontario, Canada

3. Department of Radiology, School of Medicine, 5th Azar Hospital, Radiology, Golestan University of Medical Sciences, Gorgan, Iran

4. MD, FRCPC, Department of Medical Imaging, Niagara Health System, Assistant Clinical Professor, Department of Radiology, McMaster University

5. MD, Department of Radiology, School of Medicine, 5th Azar Hospital, Golestan University of Medical Sciences, Gorgan, Golestan, Iran.

6. MD, FRCPC, Department of Medical Imaging, Niagara Health System, Associate Clinical Professor, Department of Radiology, McMaster University

7. MD, FRCPC, Department of Medical Imaging, Niagara Health System, Ontario, Canada

8. Adjunct clinical professor, department of radiology, McMaster University, Ontario, Canada

9. Center of Excellence in Precision Medicine and Digital Health, Department of Physiology, Chulalongkorn University, Thailand

## Abstract

Prostate cancer is a leading cause of death in men, yet interpreting T2-weighted prostate MRI remains challenging due to subtle, heterogeneous lesions. We developed an interpretable framework for automatic cancer detection using only 162 T2-weighted images (102 cancer, 60 normal), addressing small dataset challenges through transfer learning and data augmentation. Furthermore, we conducted the first comprehensive comparison of Vision Transformers (ViT, Swin), CNNs (ResNet18), and classical methods (Logistic Regression, SVM, HOG+SVM) on prostate MRI. Transfer-learned ResNet18 achieved the best performance (90.9% accuracy, 95.2% sensitivity, AUC 0.905) with only 11M parameters, while Vision Transformers underperformed (81–82% accuracy, 80M+ parameters). Classical HOG+SVM matched ResNet18's accuracy (AUC 0.917), proving handcrafted features can compete with deep neural networks on small datasets.

Unlike state-of-the-art models that rely on biparametric MRI (T2+DWI)—where DWI is the dominant and diagnostically simpler sequence under PI-RADS v2.1—and require large training cohorts, our lightweight approach achieves competitive AUC using only T2-weighted sequences, despite T2 being the more diagnostically challenging input, while also reducing acquisition time, artifact susceptibility, patient burden, and computational requirements.

In a reader study on 22 clinical cases, five expert radiologists achieved mean sensitivity of 67.5% with moderate inter-reader agreement (Fleiss' kappa = 0.524, p < 0.001), compared to the AI model's 95.2%, demonstrating that radiologists miss approximately 1 in 3 cancers, where the AI misses fewer than 1 in 20. These findings suggest that AI could serve as an initial screening step to help reduce missed cancers and provide more consistent results, given the variability observed between radiologists. To support transparency and reproducibility, the code, trained models, and dataset used in this study are publicly available at: https://github.com/VahidMonfared/prostate-cancer-mri-ai

## 1. Introduction

Prostate cancer remains one of the most common malignancies affecting men, and outcomes are strongly influenced by how early clinically significant disease is detected and treated. Multiparametric prostate MRI, particularly high-resolution T2-weighted imaging, has become central to contemporary risk assessment because it can localize suspicious lesions, reduce unnecessary biopsies, and better guide targeted sampling and therapy planning [1-5]. Still, real-world prostate MRI interpretation is challenging because image quality varies across MRI scanners and acquisition settings [3,14,24,25], disease can be subtle on grayscale images, and radiologist experience meaningfully affects consistency and turnaround time.

Over the past decade, prostate MRI interpretation for prostate cancer detection has moved toward standardization and quantitative decision support, with early efforts emphasizing structured reporting frameworks to reduce inter-reader variability and align MRI findings with clinical risk, including PI-RADS guidance and education-oriented overviews [1], followed by formal consensus standards and updates that refined sequence interpretation and scoring rules [2, 7]. In parallel, radiomics reframed medical images as high-dimensional quantitative data, motivating feature-based modeling beyond visual assessment alone [3]. They emphasized that once images are treated as data, a typical radiomics workflow follows: standardized image acquisition/reconstruction, careful lesion definition (segmentation), extraction of quantitative descriptors (e.g., intensity, texture, shape), and rigorous model building with proper validation [3, 26]. They also highlighted that clinical translation depends on feature robustness and reproducibility—including harmonized preprocessing, avoidance of over-fitting, and multi-site validation, so that radiomic signatures generalize beyond a single scanner, protocol, or institution.

As deep neural networks matured, broad medical-imaging surveys documented rapid gains across classification, detection, and segmentation tasks, setting the stage for prostate MRI applications [4]. Clinically, high-quality diagnostic trials reinforced the value of MRI in the diagnostic pathway (e.g., PROMIS) and supported MRI-informed strategies that improve the balance between detecting clinically significant cancer and limiting overdiagnosis [5, 6]. Building on these foundations, recent work evaluated automated and deep learning–based approaches against radiologist assessment frameworks, highlighting both the promise and the remaining gap between lab performance and dependable clinical deployment [8]. They directly benchmarked a deep learning system trained on T2-weighted and diffusion MRI against routine clinical PI-RADS assessment for identifying clinically significant prostate cancer, showing that automated algorithms can approach expert-level detection while providing consistent, reproducible outputs. At the same time, their study underscored that reliable clinical deployment hinges on robust lesion localization/segmentation performance and validation across diverse acquisition settings [3].

More recent studies [9–11] have emphasized explainability and usability, demonstrating that models can provide interpretable cues and potentially improve nonexpert performance—while also establishing benchmark diagnostic performance expectations for PI-RADS v2.1 [2] and tracking how modern architectures such as vision transformer [17]) are being used in medical image analysis. Importantly, prospective and multicenter evaluations and recent transformer-based prostate MRI methods [12, 13] further underline the trend toward clinically grounded validation rather than purely retrospective accuracy reporting .

Recent state-of-the-art approaches have obtained strong performance on prostate MRI classification, with models like Li et al. [13] reaching AUC 0.89 and Lee et al. [12] showing strong bicenter validation [20, 23,24]. However, these methods require biparametric MRI (T2+DWI) and training cohorts of 205–1,476 patients. Our study addresses a complementary challenge, which is achieving competitive performance with minimal data (162 images) using only T2-weighted sequences and lightweight architectures, making the approach more accessible for institutions with limited annotated datasets.

Nevertheless, neural network models require a large amount of data to train and cannot operate in small settings. Against the previous research works, this study focuses on a practical and often under-represented setting, small, heterogeneous T2-weighted prostate MRI datasets in common image formats (JPEG/PNG) with varying resolutions and acquisition characteristics.

This study offers a practical, end-to-end benchmark for prostate T2-MRI cancer classification that reflects how imaging is commonly available in routine care (simple 2D exports), while still using modern, interpretable AI. By comparing CNNs, transformers, and strong classical baselines side-by-side with explainability, we provide a clear and reproducible guide to help clinicians and researchers select reliable models and move toward real-world computer-aided decision support.

T2-weighted prostate MRI is difficult to interpret due to subtle and heterogeneous lesions, motivating AI-assisted diagnosis [2]. Based on obtained results, we perform a side-by-side comparison of classical machine learning baselines (SVM and logistic regression) and deep neural network approaches spanning CNNs (ResNet-style transfer learning) and transformer-based models (Vision Transformer ViT and Swin-Transformer). Furthermore, to demonstrate the generalizability of our approach, we test our model with external dataset Prostate158 [14], and report the results. Because performance alone is not enough for clinical confidence, we also incorporate Grad-CAM–based [15] visual explanations to highlight image regions that most influence model decisions, helping medical professionals quickly assess whether the model is "looking" at plausible, anatomically meaningful areas. Together, these analyses move beyond reporting a single accuracy number and instead offer a practical view of what different model families learn from T2-weighted MRI and how consistently they behave across datasets. By combining external testing with transparent visual explanations, we aim to support clinician trust and provide an evidence-based comparison of AI models for prostate cancer assessment. This work highlights which models perform most robustly on this dataset and what validation is still needed before clinical use.

All code, trained models, and the dataset used in this study are publicly available at the project repository: https://github.com/VahidMonfared/prostate-cancer-mri-ai.

## 2. Materials and Methods

### 2.1. Dataset

We assembled a small, heterogeneous dataset of T2-weighted prostate MRI slices provided as grayscale images and made this dataset publicly available to support reproducible research and benchmarking in prostate cancer detection. The dataset included 102 cancer images and 60 normal images with an approximate age range of 45–80 years, and all subjects were male.

Labels were assigned from the provided class structure (Cancer vs Normal). Figure 1 presents a prostate T2-weighted MRI image from our dataset after resolution enhancement applied for training (applied for some images). Resolution enhancement was applied to the images using Real-ESRGAN [27] (Figure 1); the remaining images were used without this enhancement.

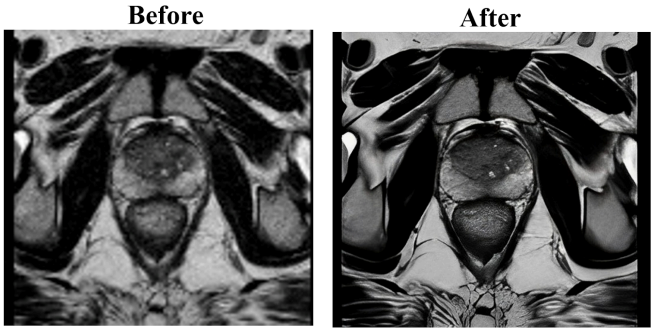

**Figure 1**. Converting low resolution image to high by Real-ESRGAN

Figure 2 shows SAM's segmentation (original Meta "Segment Anything", SAM, 2023) of the Prostate T2-MRI image, only for better understanding the regions that are needed for Grad-CAM analysis in the following next sections to realize the critical area [19].

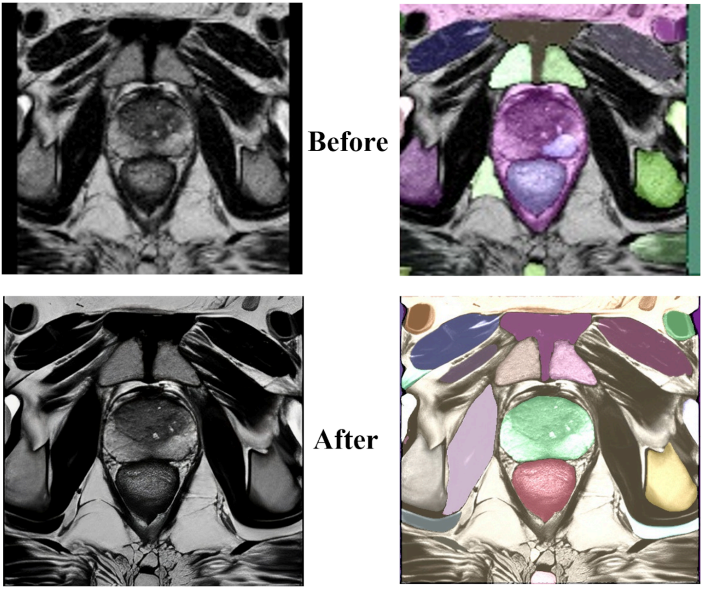

**Figure 2.** SAM segmentation of Prostate image T2-MRI with low and high resolutions

## 2.2. Study Design

We used an 80/20 stratified split to hold out a fixed test set (20%), while the remaining 80% (train+validation) was used for model development with stratified k-fold cross-validation (K=5).

This design helps reduce variance when training on small datasets while still preserving an untouched test set for final reporting.

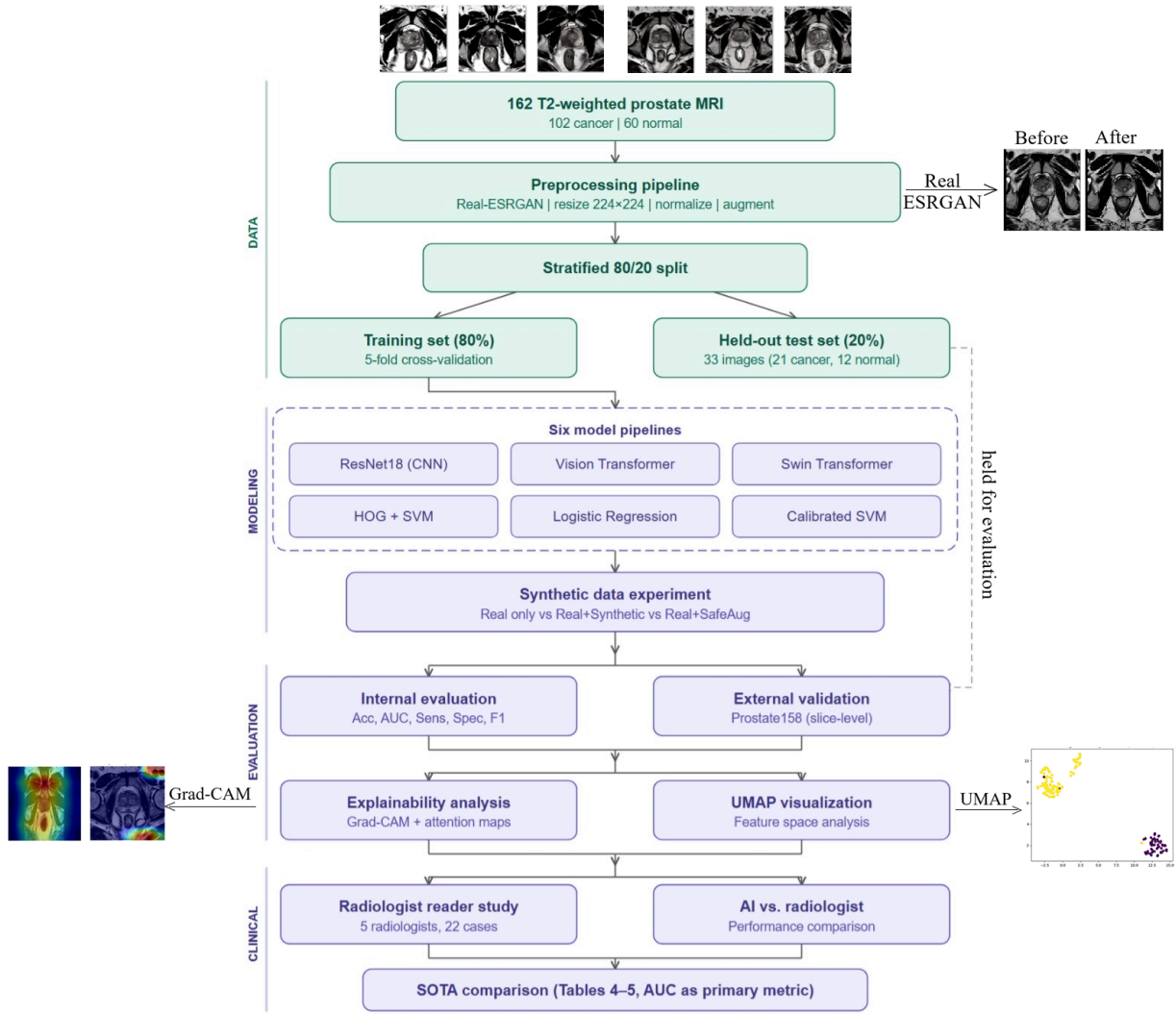

**Figure 3.** Methodology Pipeline for Prostate Cancer Detection

Figure 3 shows our complete workflow for comparing AI approaches to detect prostate cancer from MRI scans. We collected 162 T2-weighted images (102 cancer cases and 60 normal) and carefully prepared them for analysis. First, we enhanced image quality using Real-ESRGAN, Figures 1 and 2 present the before and after difference in the example above the preprocessing box. We evaluated six pipelines, including three deep neural network models (ResNet18, Vision Transformer, and Swin Transformer) [16–18], a handcrafted-feature baseline (HOG + SVM), and two traditional classifiers (logistic regression and calibrated SVM) trained on pretrained deep embeddings. Each model was trained using 5-fold cross-validation to ensure reliable evaluation, then tested on both our internal dataset and an external collection (Prostate158) [14] to check how well they generalize to new data.

To ensure a fair comparison between different modeling approaches, all models were trained using the same preprocessing pipeline, training splits, and evaluation metrics. Performance was assessed using standard classification metrics including accuracy, area under the receiver operating characteristic curve (AUC), and sensitivity. In addition to quantitative evaluation, we applied interpretability techniques such as Grad-CAM and attention visualization to better understand which image regions contributed to the model predictions. These interpretability analyses help assess whether the models focus on clinically meaningful prostate regions, which is important for building trust and transparency in AI-assisted medical imaging systems.

To understand what the models were actually focusing on, we used explainability tools like Grad-CAM; the heatmap example on the right shows how these visualizations reveal which parts of the image influence each prediction, helping doctors trust and verify the AI's decisions. In this study, we formulated a binary classification task to distinguish prostate cancer from normal findings using grayscale T2-weighted prostate MRI slices. The dataset included 102 cancer and 60 normal images, provided as variable-resolution JPEG/PNG images.

To ensure robust evaluation despite the limited dataset size, we used stratified cross-validation during training and reserved a separate hold-out test set for final performance reporting. This approach helps reduce overfitting and provides a more reliable estimate of model performance when applied to previously unseen prostate MRI data.

Our overall aim was to compare representative classical and deep neural network approaches side-by-side on the same data and report which methods are most effective and most realistic for small clinical datasets. For context, recent state-of-the-art prostate MRI AI systems are typically developed and evaluated on large, multi-institutional mpMRI datasets and benchmarks (e.g., PI-CAI) and often use volumetric 3D segmentation/detection backbones and strong task-specific pipelines; representative examples include PI-CAI challenge systems and confirmatory evaluations reported in the PI-CAI literature and related prostate lesion segmentation/detection architectures [20–22].

In contrast to these large-scale benchmark studies, our work focuses on evaluating practical AI approaches under limited-data conditions that are more typical in small clinical research settings. This allows us to assess which models remain reliable when only modest numbers of annotated prostate MRI images are available.

For model development, we first performed a stratified 80/20 train–test split, keeping 20% of the data as a fixed held-out test set. The remaining 80% of the data was then used for training and validation using stratified k-fold cross-validation (K = 5) to ensure balanced class distribution across folds. Images were standardized through a consistent preprocessing pipeline including resizing to a fixed spatial resolution (224×224 pixels), conversion from grayscale to three-channel input required by ImageNet-pretrained architectures, intensity normalization, and data augmentation (random rotation and horizontal flipping) to improve model robustness. These preprocessing strategies follow commonly used practices in deep learning–based medical image analysis pipelines [23,24].

## 2.3. Models and Training Procedures

We evaluated three deep neural network architectures: ResNet18, Vision Transformer (ViT), and Swin Transformer [4,11,16–18]. These models were trained as binary classifiers and compared using clinically meaningful metrics (accuracy, AUC, sensitivity/recall, specificity, precision, and F1-score).

In addition to reporting accuracy/AUC and related metrics, we emphasize sensitivity and specificity as primary clinically interpretable measures, and we also report model resource utilization (e.g., parameter count/model size, GPU memory footprint, and inference time per slice where applicable) to contextualize feasibility relative to heavier SOTA pipelines. Together, this design allows us to compare how different deep neural network families learn from the same T2-MRI data and to report performance in terms that align with clinical decision-making. Using AUC alongside sensitivity and specificity helps interpret not only overall accuracy, but also the trade-offs between missing cancers and generating false alarms, which is critical for evaluating potential computer-aided diagnosis tools.

We additionally evaluated classical baselines, including logistic regression, SVM using pretrained deep embeddings, and a handcrafted-feature pipeline (HOG + SVM). In addition to standard thresholding, we also evaluated threshold selection strategies that prioritize high sensitivity when the goal is to minimize missed cancers. Here, standard thresholding refers to using a fixed probability decision threshold (e.g., 0.5) without operating-point tuning; sensitivity-prioritized thresholding refers to selecting a decision threshold on the validation portion to achieve higher sensitivity when minimizing missed cancers is the priority.

To test whether "more data" helps in a small-data setting, we compared three training strategies: (i) real images only, (ii) real images plus synthetic training images, and (iii) real images with stronger (but still conservative) augmentation. The resulting performance differences (AUC, accuracy, sensitivity, and specificity) are summarized later in the Results section. The synthetic images were generated offline from the training data only by creating additional image instances through an MRI-safe transformation pipeline (e.g., small rotations/translations, mild scaling, and limited intensity/contrast perturbations). For images with low native resolution, we first applied the same resolution-enhancement step used elsewhere in preprocessing (Real-ESRGAN) [27] and then generated the synthetic variants. Importantly, all synthetic images were produced within each training fold and added only to the corresponding training split (never to validation or test) to prevent information leakage.

## 2.4. Experimental Evaluation

Finally, to evaluate generalization beyond the internal dataset, we tested the trained models on an external dataset (Prostate158 [14]) using a slice-level pipeline derived from volumetric T2-weighted images and associated lesion masks. Because prostate MRI slices may contain non-diagnostic cues (e.g., borders or labels), region-based explanations help confirm that predictions rely on prostate-region signals rather than irrelevant image areas [15]. This external evaluation provides an additional test of robustness and helps determine whether the learned representations generalize to independent data acquired under different imaging conditions.

All experiments were run with explicit random seeding (SEED=42) for Python (v3.14.3), NumPy (v2.4.2), and PyTorch (v2.7.0) where applicable. Training was configured to use GPU acceleration via NVIDIA CUDA Toolkit (v13.1 Update 1) when available. Core libraries included PyTorch/torchvision, timm, scikit-learn, and SimpleITK.

## 2.5. Radiologist Reader Study (human test)

To benchmark the performance of our approach against clinical practice, five board-certified radiologists independently reviewed 22 T2-weighted prostate MRI cases (8 clinically significant cancers confirmed by pathology, 14 unremarkable). Each radiologist identified cases they considered clinically significant cancer. Sensitivity, specificity, PPV, NPV, and accuracy were computed for each reader. Inter-reader agreement was assessed using Fleiss' kappa.

## 3. Results and Discussion

## 3.1. Overall Performance on the Internal Dataset

Across the six pipelines, models appeared to perform very well on the internal hold-out set, but their behavior differed in sensitivity–specificity tradeoffs and susceptibility to overfitting. Quantitative performance (mean ± SD across folds) is summarized in Table 1.

**Table 1.** Internal held-out test performance (20% split; image-level). The bottom row reports Mean ± SD across all six methods.

| Methods | Test Accuracy | Test AUC | Sensitivity (Recall) | Specificity | F1-Score | Confusion Matrix (TN/FP/FN/TP) | Interpretation |
|---|---|---|---|---|---|---|---|
| ResNet18 fine-tuned | **0.909** | 0.905 | **0.952** | 0.833 | **0.930** | 10 / 2 / 1 / 20 | *Strong and balanced; very few missed cancers (FN=1)* |
| Pure classic HOG + SVM | **0.909** | 0.917 | **0.952** | 0.833 | **0.930** | 10 / 2 / 1 / 20 | *Surprisingly competitive on internal data; simpler model* |
| ViT (two-stage fine-tune) | 0.818 | 0.921 | 0.810 | 0.833 | 0.850 | 10 / 2 / 4 / 17 | *Good AUC but more misses than ResNet18 (FN=4)* |
| Swin-CAM (feature-map model) | 0.818 | **0.940** | 0.762 | **0.917** | 0.842 | 11 / 1 / 5 / 16 | *Highest AUC, very high specificity, but more FN (FN=5)* |
| Pretrained features + LR (threshold=0.5) | 0.879 | **0.925** | 0.905 | 0.833 | 0.905 | 10 / 2 / 2 / 19 | *Strong baseline; interpretable and stable* |
| Pretrained features+calib SVM (threshold=0.5) | 0.818 | 0.869 | 0.905 | 0.667 | 0.864 | 8 / 4 / 2 / 19 | *High sensitivity but weaker specificity on internal test* |
| **Mean ± SD** | 0.858 ± 0.046 | 0.913 ± 0.024 | 0.881 ± 0.078 | 0.819 ± 0.082 | 0.887 ± 0.040 | — | *Summary across all six pipelines* |

**Note:** *Mean ± SD values in the bottom row summarize cross-method variability on the same held-out test set (n = 33 images: 21 cancer, 12 normal). Confusion matrix format: TN / FP / FN / TP. All metrics computed at threshold = 0.5 unless otherwise stated. Moreover LR stands for Logistic Regression, and Calib stands for Calibrated.*

When considering overall balanced performance on the internal test set (accuracy, sensitivity, and F1 score), ResNet18 and HOG+SVM performed best, achieving approximately 0.91 accuracy, 0.95 sensitivity, and F1 ≈ 0.93. In contrast, Swin-CAM achieved the highest ranking ability (AUC ≈ 0.94), but produced more false negatives (FN = 5).

A key takeaway here is that AUC alone did not guarantee the best clinical behavior on a fixed threshold. On small medical datasets, AUC can look strong even when the chosen decision threshold produces too many false negatives or false positives.

### 3.2. Parameter Sensitivity Analysis

#### 3.2.1. Cancer vs. Normal Classification at Default Probability Threshold (p = 0.5)

A fixed threshold of 0.5 provides a standardized basis for comparing all models under identical conditions. At this threshold, ResNet18 and HOG+SVM achieved the highest accuracy (0.909) with the following confusion matrix: TN=10, FP=2, FN=1, TP=20. Notably, these two models produced only one false negative (FN=1), which is clinically significant because in cancer detection, false negatives (i.e., missed cancers) represent the most consequential classification error.

#### 3.2.2. Sensitivity-Optimized Threshold Selection for Minimizing Missed Cancers

In the external-testing notebook, the model explicitly computed validation-derived thresholds aimed at very high sensitivity (min sensitivity 0.95). Lowering the decision threshold increases sensitivity at the cost of reduced specificity, consistent with the inherent sensitivity–specificity trade-off in binary classification. For example, using an FN-priority threshold (~0.0202) for the Logistic Regression model, sensitivity increased to 0.952 while specificity dropped to 0.667, resulting in more false positives but fewer missed cancers. Similarly, for the calibrated SVM model evaluated on the primary dataset, applying a sensitivity-optimized threshold of ~0.3186 achieved a sensitivity of 0.952 but reduced specificity further to 0.333, indicating a higher rate of false positives compared to the Logistic Regression model. This trade-off has direct clinical implications. In a screening context, where the primary objective is to minimize missed cancers, a lower threshold that favors higher sensitivity is generally preferred, even at the cost of increased false positives requiring further workup. Conversely, in a confirmatory or decision-support setting closer to biopsy or treatment planning, higher specificity is prioritized to reduce unnecessary invasive procedures and associated patient burden.

Therefore, for screening triage, prioritize Sensitivity (low FN) + acceptable AUC; tolerate FP. For confirmatory decision support: prioritize Specificity + PPV (low FP), while still keeping FN low. Tables 1 and 2 report full confusion matrices for each model and threshold setting, allowing direct comparison of false-negative and false-positive counts alongside summary metrics.

### 3.3. Generalization and Reliability

To assess generalization beyond the primary dataset, we evaluated our trained models on the publicly available Prostate158 dataset [14], an expert-annotated collection of 158 biparametric 3T prostate MRIs comprising T2-weighted and diffusion-weighted sequences with apparent diffusion coefficient maps. On Prostate158 (slice-level evaluation), AUCs were near chance for all neural network models we tested. External setup uses "19 cases used, 142 slices total (66 positive, 76 negative), evaluated slice-level." (see Table 2)

Table 2. External Prostate158 [14] (slice-level) performance

| Model / method | AUC | Acc | Sensitivity | Specificity | F1 | TN / FP / FN / TP | Comment |
|---|---|---|---|---|---|---|---|
| ResNet18 (baseline) | **0.494** | **0.549** | 0.500 | 0.592 | 0.508 | 45 / 31 / 33 / 33 | Best AUC among deep models, but essentially near-random discrimination |
| ResNet18 (TTA) | 0.469 | 0.542 | 0.242 | **0.803** | 0.330 | 61 / 15 / 50 / 16 | TTA increased specificity but sensitivity collapsed |
| ViT (baseline) | 0.474 | 0.486 | **0.985** | 0.053 | **0.640** | 4 / 72 / 1 / 65 | Almost always predicts "cancer" → very high sensitivity, many FPs |
| ViT (TTA) | 0.457 | 0.493 | 0.864 | 0.171 | 0.613 | 13 / 63 / 9 / 57 | Still FP-heavy; AUC not improved |
| Swin-CAM (baseline) | 0.439 | **0.549** | 0.106 | **0.934** | 0.179 | 71 / 5 / 59 / 7 | Very conservative; misses most cancers |
| Swin-CAM (TTA) | 0.416 | 0.486 | 0.879 | 0.145 | 0.614 | 11 / 65 / 8 / 58 | Flips to sensitivity-heavy; specificity collapses |

Regarding the performance of classical methods on the external Prostate158 dataset, HOG+SVM achieved an AUC of 0.277 with zero sensitivity, indicating complete failure to detect positive cases under domain shift despite high specificity. The pretrained-feature-based LR and SVM models achieved AUC values of approximately 0.45, and when sensitivity-optimized thresholds were applied, sensitivity improved to approximately 0.78 but specificity dropped to 0.13–0.22.

A key question is which approach provides the most reliable overall performance. On the internal dataset, ResNet18 and the HOG+SVM pipeline demonstrated the strongest and most stable results across the evaluated metrics. However, when evaluated on the external Prostate158 dataset, none of the approaches showed reliable generalization. The AUC values were close to chance level, and several models exhibited substantial threshold instability. For example, the Vision Transformer (ViT) tended to predict a large proportion of cases as positive, whereas Swin-CAM tended to predict a large proportion as negative, depending on the applied decision threshold and calibration.

The observed degradation in performance on Prostate158 is consistent with the well-documented domain shift problem in medical imaging, in which differences in scanner hardware, acquisition protocols, preprocessing pipelines, and labeling definitions between datasets can substantially reduce model generalization. In addition, a structural mismatch exists between the datasets used in this study: the primary dataset consists of 2D T2-weighted slices stored as JPEG/PNG images, whereas Prostate158 provides volumetric NIfTI data with voxel-level segmentation masks, further increasing the difficulty of cross-dataset generalization.

### 3.4. Explainability Analysis

Our pipeline generates Grad-CAM heatmaps and transformer attention visualizations for each prediction, enabling clinicians to verify whether model decisions are driven by anatomically plausible regions rather than image artifacts. This interpretability is particularly valuable given the near-chance external performance observed on Prostate158, as visual explanations help identify potential sources of domain shift such as off-target region focus or inconsistent intensity distributions. A detailed qualitative analysis of Grad-CAM outputs is presented in Section 3.6.

### 3.5. Impact of Synthetic Data Augmentation

To evaluate whether increasing training set size through synthetic data improves classification performance, we compared three training strategies: real images only, real images with synthetic augmentation, and real images with conservative geometric augmentation (Table 3).

Table 3. Effect of synthetic images and safe augmentation

| Training strategy | Test Accuracy | Test AUC | Sensitivity | Specificity | F1-Score | TN / FP / FN / TP | Interpretation |
|---|---|---|---|---|---|---|---|
| Real only | **0.697** | **0.861** | **0.571** | **0.917** | **0.706** | 11 / 1 / 9 / 12 | Baseline for this experiment |
| Real + Synthetic (train-only) | 0.667 | 0.810 | 0.524 | **0.917** | 0.667 | 11 / 1 / 10 / 11 | Synthetic did not improve generalization here |
| Real + SafeAug | 0.576 | 0.746 | 0.381 | **0.917** | 0.533 | 11 / 1 / 13 / 8 | Stronger augmentation hurt sensitivity the most |

The results indicate that neither synthetic augmentation nor conservative geometric transforms improved generalization performance on the held-out test set. One plausible explanation is that synthetically generated images fail to reproduce the subtle, scanner-dependent texture characteristics of real T2-weighted MRI, causing the model to learn patterns that do not transfer to authentic test images. Furthermore, even conservative augmentation strategies may distort small lesion features in 2D slices, thereby reducing model sensitivity.

The rationale for restricting augmentation to geometrically conservative transforms is grounded in the nature of MRI data: unlike natural photographs, prostate MRI slices are sensitive to intensity perturbations, and aggressive transformations such as heavy blur, color jitter, or large random crops risk eliminating clinically meaningful signals. The observed decline in sensitivity with increasing augmentation strength confirms this concern.

These findings suggest that distributional fidelity to authentic MRI characteristics is more critical than data volume in small-cohort settings, and that synthetic augmentation strategies require careful domain-specific validation before deployment in medical imaging pipelines.

### 3.6. Grad-CAM Visualization of Model Attention on Correctly Classified Test Images

Figure 4 presents Grad-CAM heatmaps for the eight cancer cases with the highest predicted probability and the eight normal cases with the lowest predicted probability (labels: y=1 cancer, y=0 normal; p = predicted cancer probability). Warmer colors (red/yellow) indicate regions with greater contribution to the model's classification decision.

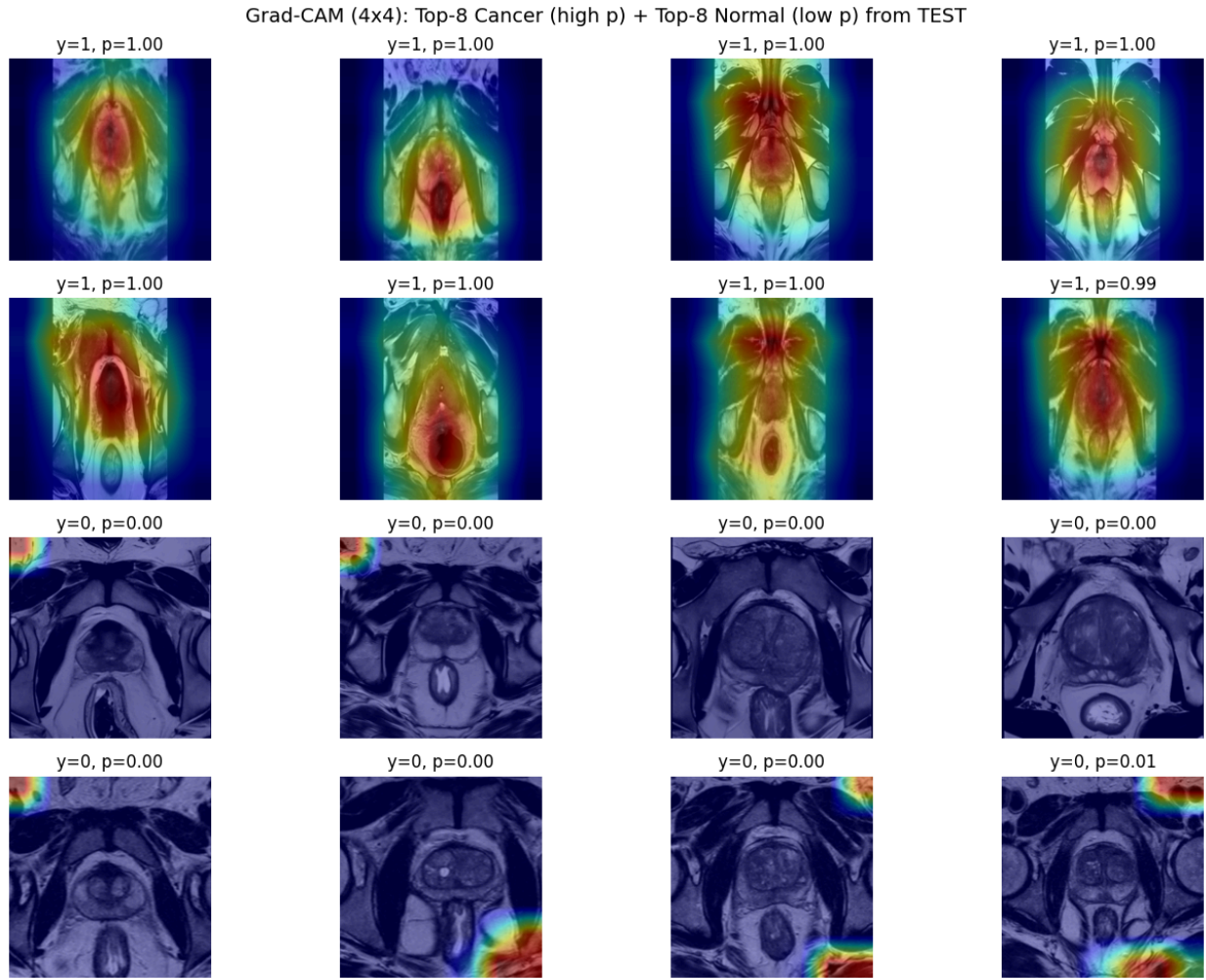

**Figure 4.** ResNet18 Grad-CAM on the held-out test set

In the cancer cases, activation was predominantly concentrated in the central prostate and pelvic region, indicating that high-confidence cancer predictions are driven by anatomically plausible structures. This spatial alignment with the expected tumor location supports the hypothesis that the model has learned clinically relevant imaging features rather than spurious correlations with image artifacts or background elements.

In the normal cases, predicted probabilities approached zero and activation maps exhibited minimal signal, consistent with the absence of suspicious findings. A small number of images displayed weak peripheral activation near image borders, likely reflecting residual sensitivity to non-diagnostic background structures rather than prostate-specific features.

These visualizations serve two practical functions in a clinical workflow. First, they enable radiologists to verify that model predictions are grounded in relevant anatomical regions, facilitating rapid quality assessment of individual predictions. Second, they provide a mechanism for identifying cases in which the model may rely on artifact-driven features, thereby supporting systematic auditing of model behavior prior to clinical integration.

### 3.7. UMAP Visualization of Learned Feature Space: ResNet18 vs. ViT

Figure 5(a–d) presents UMAP projections of the learned feature embeddings from ResNet18 and ViT, colored by true class label and predicted risk score, respectively. Both architectures produce embeddings that achieve visible separation between cancer and normal classes, though the quality and consistency of this separation differ substantially between models.

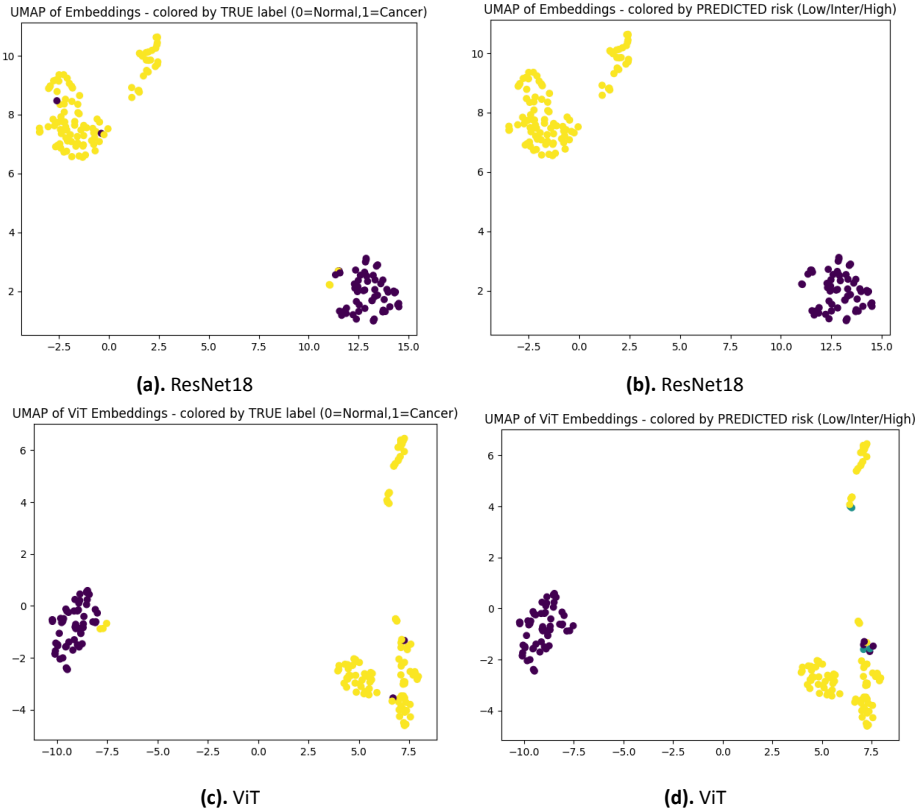

**(a).** ResNet18  **(b).** ResNet18
**(c).** ViT  **(d).** ViT

**Figure 5(a-d)**. UMAP projections of ResNet18 and ViT embeddings on the test set, colored by true label and by predicted risk, illustrate class separability and how model confidence aligns with the learned feature space.

ResNet18 embeddings exhibit compact, well-defined clusters with minimal inter-class overlap. Furthermore, the projections colored by predicted risk closely correspond to the true-label structure, indicating strong alignment between the model's confidence estimates and the underlying feature geometry. This consistency suggests that ResNet18 learns a well-calibrated representation in which decision boundaries correspond to meaningful class distinctions.

In contrast, ViT embeddings achieve class separation but with a more heterogeneous structure. The cancer class is fragmented into multiple sub-clusters, and a subset of samples near the decision boundary shows weaker correspondence between predicted risk and true label. This fragmentation likely reflects the higher data requirements of transformer architectures, which are more susceptible to representational instability when trained on limited samples with variable image characteristics.

These qualitative observations are consistent with the quantitative results reported in Table 1, where ResNet18 achieved higher accuracy and lower false-negative rates than ViT. Collectively, the UMAP analysis suggests that ResNet18 produces more stable and clinically reliable feature representations on small T2-weighted prostate MRI datasets.

### 3.8. Benchmarking Against Published Available Prostate Cancer Detection Methods

Table 4 summarizes the performance of our models alongside recent state-of-the-art methods for prostate cancer detection on MRI. Several notable comparisons emerge from this analysis.

Established approaches such as Schelb et al. [8], Hamm et al. [9], Lee et al. [12], and Li et al. [13] used cohorts ranging from 205 to 1,476 patients and relied on biparametric or multi-parametric MRI protocols, achieving AUC values of 0.79–0.89. In comparison, our transfer-learned ResNet18, trained on only 162 T2-weighted images, achieved 90.9% accuracy and an AUC of 0.905 with 11 million parameters, representing an approximately 7-fold reduction in model complexity relative to transformer-based architectures exceeding 80 million parameters.The classical HOG+SVM pipeline achieved an AUC of 0.917 with minimal computational overhead, performing comparably to Li et al.'s [13] CSwin Transformer (AUC 0.89), which required substantially larger training data and computational resources.

These results suggest that careful application of transfer learning and handcrafted feature extraction can yield clinically competitive performance even with limited training data and single-sequence imaging. This finding has practical implications for clinical settings where large annotated multi-parametric MRI datasets and dedicated GPU infrastructure are not readily available, such as community hospitals and resource-constrained environments.

**Table 4.** Comparison with Recent State-of-the-Art Models for Prostate Cancer Detection

| Study | Year | Dataset Size | Input Type | Model Type | Evaluation Level | csPCa Prevalence | AUC | Parameters |
|---|---|---|---|---|---|---|---|---|
| Schelb et al. [8] | 2019 | 312 | bp-MRI (T2+DWI) | U-Net (CNN) | Patient-level (sextant) | 42% (26/62 test) | 0.84 | NR |
| Hamm et al. [9] | 2023 | 1,224 | bp-MRI (T2+DWI) | Explainable DL (XAI) | Lesion-level | ~49% | 0.89 | NR |
| Lee et al. [12] | 2025 | 205 | bp-MRI (DLA) + mpMRI (radiologists) | Commercial DLA (Siemens) | Patient/ Lesion-level | 47% (97/205) | 0.79 | NR |
| Li et al. [13] | 2025 | 1,476 | bp-MRI (T2+DWI) | CSwin Transformer | Patient-level | ~38% (PI-CAI) | 0.89 | 80M+ |
| **Ours (ResNet18)** | **2026** | **162 images** | **T2-only** | **Transfer CNN** | **Image-level** | **64% (102/162)** | **0.905** | **11M** |
| **Ours (HOG+SVM)** | **2026** | **162 images** | **T2-only** | **Classical ML** | **Image-level** | **64% (102/162)** | **0.917** | **Minimal** |

***Note:*** *All referenced studies used biparametric MRI (T2+DWI), whereas our approach uses T2-weighted images only. Under PI-RADS v2.1, DWI is the dominant diagnostic sequence for peripheral zone lesions, where the majority of clinically significant prostate cancers arise [7]. AUC is used as the primary cross-study comparison metric because it is threshold-independent. Sensitivity, specificity, F1-score, and accuracy are reported separately in Table 5. NR = not reported.*

However, it should be noted that direct comparison across studies is limited by differences in dataset composition, imaging protocols, evaluation criteria, and clinical endpoints. AUC is used as the primary cross-study comparison metric in this work because it is threshold-independent and less sensitive to prevalence differences across datasets. Additional metrics including sensitivity, specificity, accuracy, and F1-score are reported in Table 5; however, these metrics are directly influenced by disease prevalence, which varies substantially across studies (38–64%), and should therefore be interpreted with caution rather than compared at face value.

**Table 5.** Extended Diagnostic Metrics Reported in or Derived from Referenced Studies

| Study | Evaluation Level | Prevalence | Input Type | Sensitivity | Specificity | Accuracy | F1-Score | Threshold |
|---|---|---|---|---|---|---|---|---|
| Schelb et al. [8] | Patient (sextant) | 42% (26/62) | bp-MRI (T2+DWI) | 92% | 47% | 66.1%* | 0.696* | $p \geq 0.33$ |
| Hamm et al. [9] | Lesion-level | ~49% | bp-MRI (T2+DWI) | 93% | NR | NR | NR | NR |
| Lee et al. [12] | Patient (DLA-mod.) | 47% (97/205) | bp-MRI + mpMRI | 96% | 44% | 68.8%* | 0.744* | PI-RADS $\geq 3$ |
| Li et al. [13] | Patient-level | ~38% (PI-CAI) | bp-MRI (T2+DWI) | NR | NR | NR | NR | NR |
| **Ours (ResNet18)** | **Image-level** | **64% (102/162)** | **T2-only** | **95.2%** | **83.3%** | **90.9%** | **0.930** | **$p \geq 0.5$** |
| **Ours (HOG+SVM)** | **Image-level** | **64% (102/162)** | **T2-only** | **95.2%** | **83.3%** | **90.9%** | **0.930** | **$p \geq 0.5$** |

*Note: Direct numerical comparison across rows is not appropriate due to differences in: (1) evaluation level; (2) disease prevalence, which directly affects accuracy and F1-score; (3) operating thresholds; and (4) imaging input — all referenced studies used biparametric MRI (T2+DWI), whereas our approach uses T2-weighted images only. Under PI-RADS v2.1, DWI is the dominant diagnostic sequence for peripheral zone lesions [7]. Published evidence indicates that combining T2 with DWI improves sensitivity from 51–86% to 71–89% compared with T2 alone [28]. Values marked with \* were computed from confusion matrix data reported in the original publication. NR = not reported. AUC (Table 4) remains the primary comparison metric.*

### 3.8.1. AUC and Resource Comparison Across Studies

This subsection compares AUC values, dataset sizes, and computational requirements. Models such as Li et al. [13] and Lee et al. [12] report AUC values of 0.89 and 0.79, respectively; however, these approaches operate under substantially different data and infrastructure requirements. Specifically, these methods depend on multi-parametric MRI sequences (T2-weighted, ADC, and DWI), which increase acquisition time, protocol complexity, and associated costs. Additionally, they require annotated training cohorts ranging from 205 to 1,476 patients, which limits their applicability in institutions with limited access to expert annotations. Furthermore, these architectures typically employ 80–300 million parameters, necessitating dedicated GPU infrastructure for both training and inference [20, 23, 24].

In contrast, our ResNet18 model, comprising 11 million parameters, achieved 90.9% accuracy and 95.2% sensitivity using only 162 T2-weighted images and a single MRI sequence. The HOG+SVM pipeline achieved comparable classification performance with negligible computational requirements. These results indicate that, under constrained data and resource conditions, transfer learning and handcrafted feature-based approaches can achieve diagnostic performance within a clinically relevant range of current state-of-the-art methods, while substantially reducing the barriers to implementation in resource-limited clinical environments.

**3.8.2. Practical Trade-offs and Clinical Accessibility**

The results presented above highlight a clinically relevant trade-off between model complexity and diagnostic performance. Our findings indicate that competitive classification accuracy can be achieved using a single T2-weighted MRI sequence, a small annotated dataset, and lightweight architectures, without requiring the multi-parametric imaging protocols, large-scale annotations, or high-performance computing infrastructure typical of current state-of-the-art systems. This observation is particularly relevant for community hospitals, rural healthcare facilities, and resource-constrained clinical environments where access to multi-parametric MRI, subspecialist annotations, and GPU-equipped workstations remains limited.

It should be noted that DWI is the dominant and diagnostically simpler sequence for prostate cancer detection under PI-RADS v2.1, providing high-contrast lesion visualization that substantially facilitates both human and AI-based diagnosis [7, 28]; consequently, the state-of-the-art methods compared in this study benefited from this richer imaging input, whereas our models achieved competitive or superior AUC using only T2-weighted images—the more diagnostically challenging sequence, while simultaneously reducing acquisition time, artifact susceptibility, and patient discomfort [29].

External validation on the Prostate158 dataset revealed substantial performance degradation (AUC approximately 0.45), consistent with domain shift between the primary and external datasets. While this outcome underscores the current limitations of our approach in cross-domain generalization, reporting these results transparently contributes to a more complete understanding of model behavior under realistic deployment conditions, an aspect that is frequently underreported in the literature.

Future work should investigate domain adaptation and harmonization techniques to mitigate cross-dataset performance loss while preserving the data efficiency and computational accessibility demonstrated in this study. Multi-center training with standardized preprocessing protocols and evaluation on prospectively collected cohorts will be essential steps toward clinical translation.

### 3.9. Comparison with Human Experts

Table 6 reports individual and mean diagnostic performance for five board-certified radiologists on the 22-case reader study. Inter-reader agreement was moderate (Fleiss' kappa = 0.524, 95% CI 0.314–0.734, p < 0.001), reflecting clinically meaningful variability in T2-weighted prostate MRI interpretation across readers.

**Table 6.** Radiologist Performance on 22-Case Reader Study

| Radiologist | Sens | Spec | PPV | NPV | Accuracy |
| --- | --- | --- | --- | --- | --- |
| Rad1 | 50% | 100% | 100% | 77.8% | 81.8% |
| Rad2 | 62.5% | 100% | 100% | 82.4% | 86.4% |
| Rad3 | 50% | 92.9% | 80% | 76.5% | 77.3% |
| Rad4 | 100% | 64.3% | 61.5% | 100% | 77.3% |
| Rad5 | 75% | 100% | 100% | 87.5% | 90.9% |
| **Mean** | **67.5%** | **91.4%** | **88.3%** | **84.8%** | **82.7%** |

Table 7 presents a descriptive comparison between the AI model (ResNet18) and mean radiologist performance. The AI model achieved higher sensitivity (95.2% vs. 67.5%) and overall accuracy (90.9% vs. 82.7%), whereas the radiologists demonstrated higher mean specificity (91.4% vs. 83.3%). These complementary performance profiles suggest a potential sequential workflow in which the AI model functions as a high-sensitivity initial reader to reduce the rate of missed cancers, followed by radiologist review of flagged cases to improve specificity and minimize unnecessary biopsies. It should be noted that the AI model and radiologists were evaluated on non-overlapping case sets; consequently, this comparison is descriptive rather than statistically paired. Future studies should evaluate both on identical cases to enable formal paired statistical testing.

From a deployment perspective, ResNet18 presents several practical advantages over transformer-based architectures for clinical integration. With 11 million parameters, compared to over 80 million in Vision Transformers, ResNet18 supports inference on standard CPU hardware without requiring dedicated GPU infrastructure. Through transfer learning, ResNet18 achieved 90.9% accuracy with only 162 training images, whereas transformer architectures typically require substantially larger datasets to achieve comparable performance. Among all models evaluated, ResNet18 achieved the highest sensitivity (95.2%), corresponding to a single missed cancer case compared to four to five missed cases for transformer models (Table 1).

Table 7. AI (ResNet18) vs Mean Radiologist Performance

| Metric | AI (ResNet18) | Radiologists |
|---|---|---|
| Sensitivity | 95.2% | 67.5% |
| Specificity | 83.3% | 91.4% |
| PPV | 90.9% | 88.3% |
| NPV | 90.9% | 84.8% |
| Accuracy | 90.9% | 82.7% |

Additionally, ResNet18 is fully compatible with Grad-CAM, producing spatially precise heatmaps that offer more localized interpretability than transformer attention maps. Inference time was below 100 milliseconds per image, supporting real-time integration into clinical reading workflows. Collectively, these properties position ResNet18 as a suitable candidate for deployment in clinical environments where large annotated datasets, subspecialty radiological expertise, and high-performance computing infrastructure are not readily available.

## 4. Conclusions

Among all evaluated models, transfer-learned ResNet18 and the classical HOG+SVM pipeline achieved the highest overall performance on the primary dataset, with 90.9% accuracy, 95.2% sensitivity, and an AUC of 0.905. At the default classification threshold, ResNet18 produced only one false negative out of 21 cancer cases. Grad-CAM visualizations confirmed that high-confidence predictions were predominantly driven by activation within the prostate and pelvic region, supporting anatomically plausible decision-making.

Synthetic data augmentation did not improve generalization performance, suggesting that distributional fidelity to authentic MRI characteristics is more critical than increasing training set volume in small-cohort settings. In comparison with state-of-the-art methods that utilize biparametric MRI protocols (T2+DWI) with training cohorts of 205–1,476 patients and architectures exceeding 80 million parameters, ResNet18 achieved competitive results with 11 million parameters and 162 T2-weighted images, enabling deployment on standard hardware without dedicated GPU infrastructure. Furthermore, this performance was achieved using only T2-weighted images—the more diagnostically challenging MRI sequence—without the benefit of DWI, which is the dominant and diagnostically simpler sequence for peripheral zone lesions under PI-RADS v2.1 [7, 28], while additionally eliminating the longer acquisition time, artifact susceptibility, and patient discomfort associated with DWI protocols [29].

A reader study involving five board-certified radiologists demonstrated moderate inter-reader agreement (Fleiss' kappa = 0.524, p < 0.001) and a mean sensitivity of 67.5%, compared with the AI model's sensitivity of 95.2%. This difference indicates that subjective MRI interpretation is associated with substantial diagnostic variability, whereas the AI model provides consistent and reproducible predictions. These findings support the potential utility of a sequential triage workflow in which the AI serves as a high-sensitivity initial screening tool, followed by radiologist review of flagged cases to optimize specificity and reduce unnecessary invasive procedures.

External validation on the Prostate158 dataset [14] revealed performance degradation consistent with domain shift (AUC approximately 0.49). Additionally, the AI and radiologist evaluations were conducted on non-overlapping case sets, precluding paired statistical comparison. Future work should prioritize multi-center training with harmonized preprocessing protocols, concurrent evaluation of AI and radiologists on identical case sets to enable formal statistical testing, and prospective clinical trials to validate the proposed triage workflow in routine practice.

## References

[1] T. Barrett, B. Turkbey, and P.L. Choyke, "PI-RADS version 2: what you need to know," Volume 70, Issue 11, November 2015, Pages 1165-1176. doi: 10.1016/j.crad.2015.06.093 (https://doi.org/10.1016/j.crad.2015.06.093).

[2] J. C. Weinreb et al., "PI-RADS Prostate Imaging–Reporting and Data System: 2015, Version 2," European Urology, vol. 69, no. 1, pp. 16–40, 2016. doi: 10.1016/j.eururo.2015.08.052 (https://doi.org/10.1016/j.eururo.2015.08.052).